\documentclass[letterpaper, 10 pt, conference]{ieeeconf}  %

\IEEEoverridecommandlockouts                              %

\usepackage{cite}
\usepackage{mathdots}
\usepackage{amsmath,amssymb,amsfonts}
\usepackage{textcomp}
\def\BibTeX{{\rm B\kern-.05em{\sc i\kern-.025em b}\kern-.08em
    T\kern-.1667em\lower.7ex\hbox{E}\kern-.125emX}}

\usepackage[colorlinks=True,allcolors=blue]{hyperref}
\usepackage[margin=1in]{geometry}

\usepackage{wrapfig}

\usepackage{graphicx}
\usepackage{amsmath}
\usepackage{amssymb}
\usepackage{bm}
\usepackage{accents}

\newcommand{\uwidehat}[1]{%
  \mathpalette\douwidehat{#1}%
}
\makeatletter
\newcommand{\douwidehat}[2]{%
  \sbox0{$\m@th#1\widehat{\hphantom{#2}}$}%
  \sbox2{$\m@th#1x$}
  \sbox4{$\m@th#1#2$}
  \dimen0=\ht0
  \advance\dimen0 -.8\ht2
  \dimen2=\dp4
  \rlap{%
    \raisebox{\dimexpr\dimen0-\dimen2}{%
      \scalebox{1}[-1]{\box0}%
    }%
  }%
  {#2}%
}
\makeatother

\def\namedlabel#1#2{\begingroup
    #2%
    \def\@currentlabel{#2}%
    \phantomsection\label{#1}\endgroup
}
\makeatletter
\newcommand{\leqnomode}{\tagsleft@true\let\veqno\@@leqno}
\makeatother

\makeatletter
\newcommand{\customlabel}[2]{%
\protected@write\@auxout{}{\string \newlabel{#1}{{#2}{\thepage}{#2}{#1}{}}}%
\hypertarget{#1}{#2}
}
\makeatother

\newcommand\PMPODE[1]{\hyperref[PMPODE]{$\textbf{ODE}_{#1}$}\xspace}
\newcommand\PMPODElambda[1]{\hyperref[PMPODElambda]{$\textbf{ODE}_{#1}^\lambda$}\xspace}
\newcommand\PMPODElambdaunder[1]{\hyperref[PMPODElambda_under]{$\uwidehat{\textbf{ODE}_{#1}^\lambda}$}\xspace}
\newcommand\PMPODElambdaover[1]{\hyperref[PMPODElambda_over]{$\widehat{\textbf{ODE}_{#1}^\lambda}$}\xspace}
\newcommand\PMPODEepsilonfullrank[1]{\hyperref[PMPODEepsilon_full_rank]{$\textbf{ODE}_{#1}^\epsilon$}\xspace}

\newcommand\PMPODEg[1]{\hyperref[PMPODEg]{$\textbf{ODE}^g_{#1}$}\xspace}
\newcommand\OCPg[1]{\hyperref[OCPg]{$\textbf{OCP}^g_{#1}$}\xspace}

\newcommand\OCP[1]{\hyperref[OCP]{$\textbf{OCP}_{#1}$}\xspace}
\newcommand\BVP[1]{\hyperref[BVP]{$\textbf{BVP}_{#1}$}\xspace}
\newcommand\OCPMPC[1]{\hyperref[OCPMPC]{$\textbf{OCP}(#1)$}\xspace}

\usepackage{tabularx}
\newcolumntype{C}{>{\centering\arraybackslash}p{19mm}}
\newcolumntype{G}{>{\centering\arraybackslash}p{4mm}}
\newcolumntype{S}{>{\centering\arraybackslash\scriptsize}p{4mm}}

\usepackage[font=small,labelfont=bf]{caption}
\usepackage{color}
\usepackage[usenames,dvipsnames]{xcolor}

\usepackage{cleveref} %

\usepackage{xspace}

\let\labelindent\relax
\usepackage{enumitem} %

\usepackage{cite} %

\usepackage{amssymb}%
\usepackage[plain]{algorithm}
\makeatletter
\renewcommand*{\ALG@name}{Alg.}
\makeatother
\usepackage[noend]{algpseudocode}

\usepackage{xpatch} %

\usepackage{tabularx, booktabs} %
\newcolumntype{F}{>{\centering\arraybackslash}p{1.9cm}}

\newcolumntype{G}{>{\centering\arraybackslash}p{1.2cm}}
\newcolumntype{H}{>{\centering\arraybackslash}p{0.4cm}}
\newcolumntype{I}{>{\centering\arraybackslash}p{0.92cm}}
\newcolumntype{Z}{>{\centering\arraybackslash}p{-0.92cm}}

\newtheorem{remark}{Remark}

\newcommand\mydots{\hbox to 1em{.\hss.\hss.}}

\newcommand{\dd}{\textrm{d}}

\newcommand{\E}{\mathbb{E}}
\newcommand{\R}{\mathbb{R}}

\newtheorem{preremark3}{Theorem}[section]

\usepackage{mdframed}
\usepackage{lipsum}
\newmdtheoremenv{theo}{Theorem}

\usepackage{multirow}

\usepackage[many]{tcolorbox}
\tcbuselibrary{breakable}
\makeatletter
\def\tcb@parbox@use@false{%
  \def\@parboxrestore{\linewidth\hsize\let\@parboxrestore=\tcb@parboxrestore}%
}
\makeatother

\title{\LARGE \bf
Risk-Averse Model Predictive Control\\for Racing in Adverse Conditions
}

    \author{Thomas Lew$^{1}$,
Marcus Greiff$^{1}$,
Franck Djeumou$^{1,2}$,\\
Makoto Suminaka$^{1}$,
Michael Thompson$^{1}$,
John Subosits$^{1}$
    \thanks{$^{1}$  Toyota Research Institute, Los Altos, CA, USA}
\thanks{$^{2}$ %
Rensselaer Polytechnic Institute, Troy, NY, USA}
    }

\begin{document}

\maketitle
\thispagestyle{empty}
\pagestyle{empty}

\begin{abstract}
Model predictive control (MPC) algorithms can be sensitive to model mismatch when used in challenging nonlinear control tasks. In particular, the performance of MPC 
for vehicle control at the limits of handling suffers when the underlying model overestimates the vehicle's performance capability. 
In this work, we propose a risk-averse MPC framework that explicitly accounts for uncertainty over friction limits and  tire parameters. %
Our approach leverages a sample-based approximation of an optimal control problem with a conditional value at risk (CVaR) constraint. 
This sample-based formulation enables planning %
with a set of expressive vehicle dynamics models using different tire parameters. 
Moreover, this formulation %
enables efficient numerical resolution 
 via sequential quadratic programming and GPU parallelization. 
Experiments on  a Lexus LC 500 show that risk-averse MPC unlocks reliable performance, 
while a deterministic baseline that plans using a single dynamics model may lose control of the vehicle in adverse road conditions. %
\end{abstract}

\section{Introduction}
Expert racing drivers are able to pilot a vehicle at its performance limits by using all the available friction potential between the tires and the road. 
They are able to do this reliably lap after lap despite changes in the vehicle's performance and behavior due to tire temperature, tire wear, and especially, weather conditions. 
However, current approaches to autonomous vehicle control struggle in such settings because they are sensitive to discrepancies between the model used for control and the true system \cite{weber2023modeling}. %
This sensitivity motivates the design of new algorithms 
that can robustly leverage the full vehicle capabilities.  
Designing reliable control algorithms for racing may inform the future design of expert driver assistance systems by unlocking reliable responses for avoiding sudden obstacles, driving in adverse weather, and reacting quickly to challenges on the road.

Research in autonomous racing has boomed in the last decade, see  \cite{Betz2022} for a survey. 
State of the art control approaches to racing use model predictive control (MPC) 
to maximize path progress along a planning horizon while respecting constraints such as track bounds. These works use vehicle dynamics models of varying fidelity such as point mass models \cite{Subosits2019,Wischnewski2023}, singletrack models with Pacejka \cite{Liniger2014} and Fiala tire models \cite{Dallas2023,Weber2024}, and data-driven models \cite{Kabzan2019}. 
However, vehicle dynamics models have limitations, and although perceiving the environment to predict changes in road conditions and online adaptation can help reduce model mismatch, some factors like black ice may be difficult to observe, while reactive online learning methods may not be fast-enough to avoid  leading the vehicle into an unrecoverable state.

\begin{figure}[!t]
    \centering
      \includegraphics[width=1.0\linewidth]{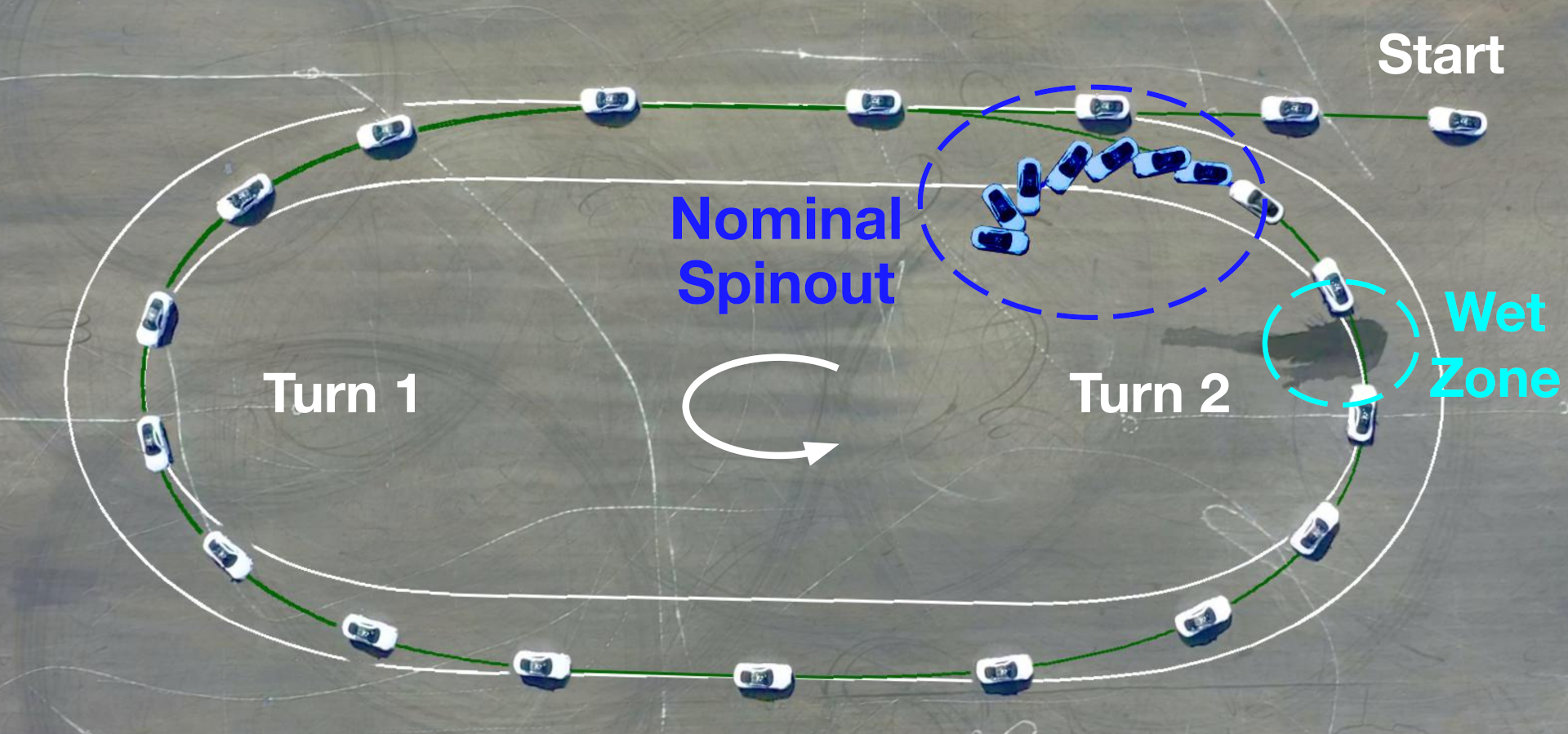}
            \caption{Racing results through a wet area. The proposed risk-averse MPC reliably handles the vehicle throughout the turn. In contrast, a deterministic MPC controller consistently spins out the vehicle as it over-predicts the attainable tire forces.}
            \vspace*{-8mm}
    \label{fig:hardware:racing_wet:aerial}
            
\end{figure} 

These modeling challenges motivate %
the design of MPC tools that explicitly account for uncertainty %
to optimally trade off robustness and performance.  
Previous uncertainty-aware MPC methods for racing use stochastic MPC \cite{Liniger2017} and tube MPC 
\cite{Wischnewski2023}. 
Using a linear model of the vehicle with additive disturbances capturing model mismatch, these methods have demonstrated reliable racing performance. In \cite{Kabzan2019}, using Gaussian process models %
allowed online adaptation to gradually improve laptime. 
However, for computational reasons, this method fixes the uncertainty estimates during the optimization, so the MPC is not incentivized to select actions that may reduce uncertainty and lead to faster racing despite uncertainty. %
Also, these works use a linearization-based uncertainty propagation scheme or model uncertainties with additive disturbances that are randomized at each time, and thus  neglect correlations over time. 
However, uncertain tire friction properties (e.g., due to driving over a wet road) may be highly correlated over time. Treating such sources of uncertain model mismatch %
as random disturbances may lead to suboptimal racing or to under-estimating the likelihood of spinning out.
As shown in \cite{Mordatch2015,Abraham2020,Brudigam2021,Dyro2021,LewExact2023,Prajapat2024,LewBonalliPavoneRAL2024}, sample-based uncertainty-aware optimization methods offer a potential avenue for using nonlinear dynamics models and accounting for complicated sources of uncertainties, which may necessary for expert racing.

\textbf{Contributions}: We introduce a risk-averse  approach for racing that explicitly reasons over uncertainty: 
\begin{itemize}[leftmargin=3.5mm]
\item We propose a risk-constrained racing formulation that explicitly accounts for uncertainty over tire forces. It includes 
a conditional value at risk (CVaR) constraint for track bounds, nonlinear uncertain dynamics, and a cost to minimize expected lap time. 
\item We reformulate the risk-constrained problem using samples of the tire model parameters. %
Then, we propose a numerical optimization scheme that leverages the sparsity of the problem and parallelizes computations on a graphics processing unit (GPU), unlocking an uncertainty-aware MPC scheme that reasons over $10$ different dynamics parameters and runs at $20$Hz.
\end{itemize}
We validate the controller on a Lexus LC 500 and show that the  proposed risk-constrained MPC approach %
unlocks reliable performance, while deterministic MPC may lose control of the vehicle in adverse conditions. 

\section{Vehicle Dynamics and Tire Model}
\begin{figure}[!t]
    \centering
    \includegraphics[width=1.0\linewidth]{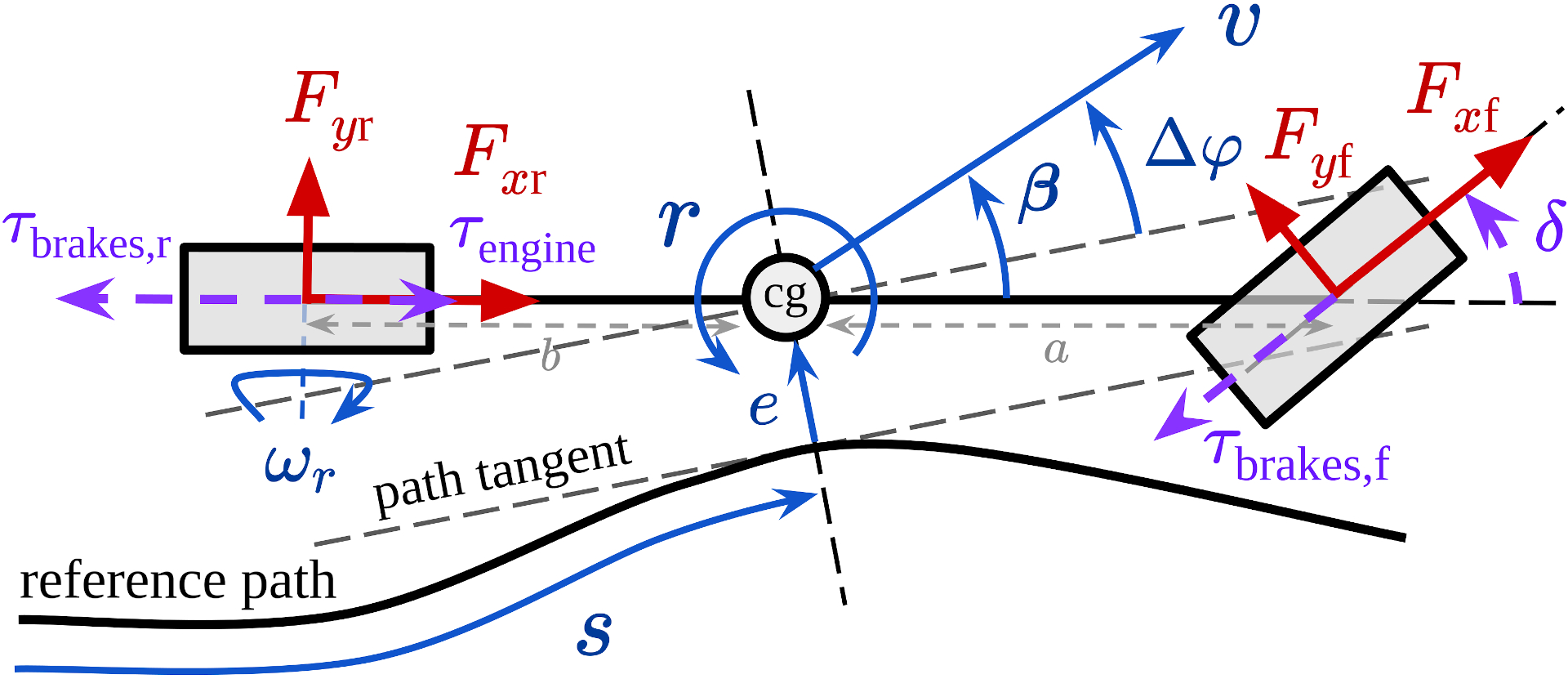}
    \caption{Vehicle on the reference path.}
    \label{fig:states}
            \vspace{-3mm}
\end{figure}

We use a single-track vehicle model in curvilinear coordinates \cite{Goh2019,Dallas2023,Kobayashi2024}, see Figure \ref{fig:states}. We model wheelspeed dynamics to better account for wheel slippage and load transfer dynamics to account for variation in tire normal forces. 
We define states and control inputs
\begin{align*}
\tilde{x}(t)&=
(
r(t), 
v(t),
\beta(t),
\omega_r(t),
\Delta F_z(t),
e(t),
\Delta\varphi(t),
s(t)
)
\\
\tilde{u}(t)&=
(
\delta(t), %
\tau_{\textrm{engine}}(t),
\tau_{{\textrm{brakes},f}}(t),
\tau_{\textrm{brakes},r}(t)
),
\end{align*}
where $r$ is the yaw rate, $v$ is the total velocity, $\beta$ is the sideslip, $\omega_r$ is the rear wheelspeed, $\Delta F_z$ is the load transferred from the front to rear axle, $e$ and $\Delta\varphi$ are the lateral and angle deviations to the reference trajectory, $s$ is the progress along the reference trajectory, $\delta$ is the steering angle, $\tau_{\textrm{engine}}\geq 0$ is the engine torque, and $\tau_{{\textrm{brakes},f}},
\tau_{\textrm{brakes},r}\leq 0$ are the front and rear brake torques, respectively. 
We model the vehicle dynamics  as
\vspace*{-3mm}

{ 
\small
\begin{equation}\label{eq:dynamics:in_time}
\begin{bmatrix}
\dot{r}
\\[1mm]
\dot{v}
\\
\\
\dot{\beta}
\\
\\
\dot{\omega}_r
\\[1mm]
\Delta \dot{F}_z
\\[1mm]
\dot{e}
\\
\Delta\dot{\varphi}
\\
\dot{s}
\end{bmatrix}
=
\begin{bmatrix}
(
a
	F_{yf}\cos\delta
+
	aF_{xf}\sin\delta
 - 
b F_{yr})/I_z
\\
\big(-F_{yf}\sin(\delta-\beta)+F_{xf}\cos(\delta-\beta)
+
\\
\hspace{20mm}
F_{yr}\sin\beta
+
F_{xr}\cos\beta
\big)/m
\\
-r +
\big(
F_{yf}\cos(\delta-\beta)+F_{xf}\sin(\delta-\beta)+
\\
\hspace{20mm}
F_{yr}\cos\beta-F_{xr}\sin\beta\big)/(mv)
\\
(\tau_{\textrm{engine}}
+
\tau_{\textrm{brakes},r}
-F_{xr}r_\textrm{w})/I_\textrm{w}
\\
-c(\Delta F_z -\frac{h_{\textrm{cg}}
}{a+b}(F_{xr} + F_{xf} \cos\delta- F_{yf}\sin\delta
)
)
\\
v\sin(\Delta\varphi)
\\
\dot\beta+\dot{r}-\kappa_{\textrm{ref}}v\cos(\Delta\varphi)/(1-\kappa_{\textrm{ref}}e)
\\
v\cos(\Delta\varphi)/(1-\kappa_{\textrm{ref}}e)
\end{bmatrix}
\end{equation}
}%
where 
$(a,b)$ are the distances from the center of gravity to the front and rear axles, 
$(I_z,m)$ are the vehicle inertia and mass, $(r_\textrm{w},I_\textrm{w})$ are the wheel radius and the rear axle inertia, 
$c>0$ is a constant, 
$h_{\textrm{cg}}$ is the center of gravity height, 
$\kappa_{\textrm{ref}}$ is the curvature of the reference path. %

To model tire forces $(F_{xf},F_{yf},F_{xr},F_{yr})$, we use the isotropic coupled slip brush Fiala model \cite{fiala1954seitenkraften,svendenius2007tire}
\begin{align*}
F_{yf} &= -F_{\textrm{total},f}\tan(\alpha_f)/\sigma_f, 
&&F_{xf} = r_\textrm{w}\tau_{{\textrm{brakes},f}}, 
\\
F_{yr} &= -F_{\textrm{total},r}\tan(\alpha_r)/\sigma_r,
&&F_{xr} = F_{\textrm{total},r}\kappa_r/\sigma_r.
\label{eq:forces}
\end{align*}
The magnitude of the tire forces $(F_{\textrm{total},f},F_{\textrm{total},r})$ are
\begin{align*}
F_\textrm{total} &=  
\begin{cases}
C\sigma - \frac{C^2 \sigma^2}{3F_\textrm{max}}  + \frac{C^3\sigma^3}{27(F_\textrm{max})^2} &\text{if } |\sigma| < \sigma_\textrm{slip}\\
F_\textrm{max} &\text{if } |\sigma| \geq \sigma_\textrm{slip}\\
\end{cases},
\end{align*}
where 
$(\sigma_f,\sigma_r)$ are the total tire slips, and $(\sigma_{\textrm{slip},f},\sigma_{\textrm{slip},r})$ are the total slips as the tires begin fully sliding
\begin{align*}
\sigma = \sqrt{\tan(\alpha)^2 + \kappa^2},
\quad
\sigma_\textrm{slip}=\tan^{-1}(3\mu F_z/ C).
\end{align*}
The tire loads $(F_{zf},F_{zr})$ depend on the static tire loads $(F_{\textrm{nom},zf}, F_{\textrm{nom},zr})$ as $F_{zf}=F_{\textrm{nom},zf}-\Delta F_z$ and $F_{zr}=F_{\textrm{nom},zr}+\Delta F_z$. 
The maximal tire forces $F_{\textrm{max}}$ are
$$
F_{\textrm{max},f}=\sqrt{(\mu F_{zf})^2-(r_\textrm{w}\tau_{{\textrm{brakes},f}})^2}.
\quad
F_{\textrm{max},r}=\mu F_{zr},
$$
The lateral slip angles $(\alpha_f,\alpha_r)$ and longitudinal slip ratios $(\kappa_f,\kappa_r)$ are 
\begin{align*}
\tan(\alpha_f) &= (v\sin\beta+ar)/(v\cos\beta)-\delta,
\ \ 
\kappa_f=0,
\\
\tan(\alpha_r)&=(v\sin\beta-br)/(v\cos\beta), \:
\kappa_r= \frac{(r_{\textrm{w}}\omega_r-v)}{v}.
\end{align*}
The rear tire model explicitly computes $\kappa_r$ since the rear wheelspeed $\omega_r$ is in the vehicle state. The front tire model captures coupled lateral-longitudinal tire forces by derating the $F_{\textrm{max}}$ term according to the friction circle. 

\begin{figure}[!t]
    \centering
    \includegraphics[width=1.0\linewidth]{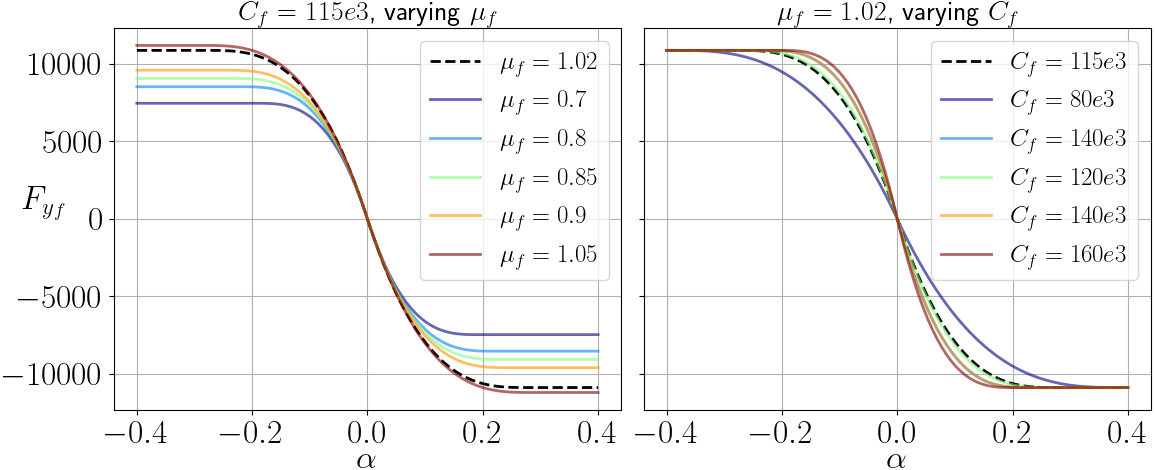}
    \caption{Tire forces $F_{yf}$ as a function of the slip angle $\alpha$.
\vspace{-10pt}
}
    \label{fig:tire_forces}
\end{figure}

Many environmental factors can influence tire behavior including road conditions, tire condition, and even tire temperature \cite{Kobayashi2024}.  
With our tire model,  the tire forces $(F_{xf},F_{yf},F_{xr},F_{yr})$ depend on the tire-road friction $\mu$ and the tire stiffness $C$ parameters, hereinafter defined
\begin{equation}\label{eq:tire_parameters}
\theta=(\mu_f,\mu_r,C_f,C_r).
\end{equation}
In Figure \ref{fig:tire_forces}, 
we represent the front tire lateral forces $F_{yf}$ as a function of the slip angle $\alpha$ for different values of  $(\mu_f,C_f)$. We observe that tire forces are sensitive to the choice of these parameters. Inaccurate parameter estimation may cause over-estimation of the maximum tire forces. This motivates accounting for uncertainty over these parameters to design a reliable controller.

\section{Minimum-Time Formulation for Racing}
Next, we describe
the racing formulation used in this work. We reparameterize the state and control input as a function of path progress, add the time variable to the state, and combine the rear braking and engine torques to avoid simultaneous acceleration and braking without the need for a non-convex constraint or cost. %
We define
\begin{align*}
x(s)&=
(
r(s), 
v(s),
\beta(s),
\omega_r(s),
\Delta F_z(s),
e(s),
\Delta\varphi(s),
t(s)
)
\\
u(s)&=
(
\delta(s), %
\tau_{\textrm{combined},r}(s),
\tau_{\textrm{brakes},f}(s)%
),
\end{align*}
and map the combined rear torque  to the engine and rear brake torques via $\tau_{\textrm{engine}}=\max(0, \tau_{\textrm{combined},r})$ and $\tau_{\textrm{brakes},r}=\min(0, \tau_{\textrm{combined},r})$. 
The dynamics $x'(s):=\frac{\dd x(s)}{\dd s}$ of the resulting system are
\begin{equation}\label{eq:dynamics}
x'(s)=
\frac{1}{\dot{s}}\left(
\dot{r},\dot{v},\dot{\beta},\dot{\omega}_r,
\Delta \dot{F}_z,
\dot{e},\Delta\dot{\varphi},1
\right)(s),
\end{equation}
where $(\dot{r},\dot{v},\dot{\beta},\dot{\omega}_r,
\Delta \dot{F}_z,
\dot{e},\Delta\dot{\varphi},\dot{s})$ are given in \eqref{eq:dynamics:in_time}. 

We discretize the dynamics with a constant path progress difference $\Delta s=3\,\textrm{m}$ and a trapezoidal scheme
\begin{equation}\label{eq:trapezoidal}
x_{k+1}
=
x_k+\frac{\Delta s}{2}(x'_k+x'_{k+1}),
\end{equation}
where  %
$(x_k,u_k)\approx (x(k\Delta s),u(k\Delta s))$ 
correspond to the state and control at $s=k\Delta s$ along the path (assuming that $x_0$ corresponds to $s=0$ without loss of generality). 
Using a trapezoidal scheme 
for the dynamics constraints \eqref{eq:trapezoidal} allows us to plan with a coarse discretization $\Delta s$ over large distances $N\Delta s$. We found that using explicit integrators instead (e.g., an Euler or high-order Runge Kutta schemes) can cause numerical instability if the minimum time cost dominates (see \eqref{eq:cost}). %
To highlight the dependency on the tire parameters $\theta=(\mu_f,\mu_r,C_f,C_r)$, the dynamics constraint \eqref{eq:trapezoidal} are equivalently written as
\begin{equation}\tag{\ref{eq:trapezoidal}}
f(x_{k+1},x_k,u_k,u_{k+1},\theta)=0.
\end{equation}
The racing objective consists of minimizing lap time. 
To improve the robustness  and smoothness of the %
controller, we also penalize state-control deviations to the reference $(x_{\textrm{ref}},u_{\textrm{ref}})$ %
and fast changes in the control inputs. We define the linear and quadratic costs
\begin{align}
\label{eq:cost}
\ell_T(x)&=t,
\ \,  \ell(x,u,w)
=(x-x_{\textrm{ref}})^\top Q(x-x_{\textrm{ref}})
\,+
\hspace{-2mm}
\\
&\hspace{-7mm} 
(u-u_{\textrm{ref}})^\top R(u-u_{\textrm{ref}})
\,+
(u-w)^\top W(u-w)/(\Delta s^2),
\nonumber
\end{align}
where $(Q,R,W)$ are diagonal matrices with small entries compared to the terminal time cost $\ell_T$. 
Constraints account for actuator limits $(u_{\textrm{min}},u_{\textrm{max}})$ and ensure that the vehicle remains within the track bounds $(e_{\textrm{min}},e_{\textrm{max} })$. To plan over a horizon $N$ from an initial state and control, %
we formulate the optimal control problem (\textbf{OCP})
\begin{align*}
\min_{x,u}
\ 
&\ell_T(x_N)+%
\sum_{k=0}^{N-1}
\ell(x_k,u_k,u_{k+1})
&&
\hspace{11mm}\boxed{\scalebox{1.1}{\textbf{OCP}}}
\\
\text{s.t.}
\ 
&f(x_{k+1},x_k,u_k,u_{k+1},\theta)=0, &&k\in[0,N-1],
\\
&(x_0,u_0)=(x_{\text{init}},u_{\text{init}}),
\\
&u_{\textrm{min}}\leq u_k\leq u_{\textrm{max}},
&&k\in[1,N],
\\
&e_{\textrm{min},k}\leq e_k \leq e_{\textrm{max},k},
&&k\in[1,N],
\end{align*}
with the notation $[0,N]:=\{0,\dots,N\}$. 
The only non-convex term in \textbf{OCP} is the dynamics constraints. %

\textit{MPC \& linear interpolation}: 
For real-time control, \textbf{OCP} is solved recursively from the current $(x_{\text{init}},u_{\text{init}})$ and the plan $(u_0,\dots,u_N)$ is sent to the vehicle. A low-level controller executes the control $u(s)$ at the current $s$ via linear interpolation of the latest plan $(u_0,\dots,u_N)$.

\section{Risk-Averse Formulation}
Solutions to \textbf{OCP} may be sensitive to the value of the tire parameters $\theta$, and a controller that uses inaccurate values of $\theta$ may result in poor closed-loop performance. Intuitively, low friction values should result in smaller torque values to avoid spinning out or violating track bounds. In this section, we formulate a risk-averse MPC problem that accounts for uncertainty over the parameters $\theta$ distributed according to a probability distribution $p_\theta$.  
First, we reformulate the track bound constraints $e_{\textrm{min},k}\leq e_k \leq e_{\textrm{max},k}$ in \textbf{OCP}  as
\begin{equation}\label{eq:constraint:g}
g_k(x_k):=|e_k-e_{\textrm{mid},k}|-\Delta  e_k \leq 0,
\end{equation}
with $
e_{\textrm{mid},k}=(e_{\textrm{max},k}+e_{\textrm{min},k})/2$ and $\Delta e_k=\left|e_{\textrm{max},k}-e_{\textrm{min},k}\right|/2
$. This reformulation combines the two constraints $e_{\textrm{min},k}\leq e_k$ and $e_k \leq e_{\textrm{max},k}$ into one, simplifying the latter formulation of the risk constraint. The absolute value $|e_k-e_{\textrm{mid},k}|$ in \eqref{eq:constraint:g} does not affect numerical stability, since only one of the two constraints $e_{\textrm{min},k}\leq e_k$ or $e_k \leq e_{\textrm{max},k}$ can be active at a solution. 

To enforce track bound constraints and account for uncertainty over state trajectories due to uncertain tire forces,  we constrain %
the tail probability of ever violating the constraint $g_k(x_k)\leq 0$ in  \eqref{eq:constraint:g} at \textit{any} timestep $k$  
\begin{equation}\label{eq:avar:constraint}
\textrm{CVaR}_\alpha\left(
\max_{k=1,\dots,N}g_k(x_k)\right)\leq 0,
\end{equation}
where 
$\textrm{CVaR}_\alpha$ is the conditional value at risk at level $\alpha\in(0,1)$ (or average value at risk \cite{Shapiro2014,LewBonalliPavoneRAL2024}), defined as $\textrm{CVaR}_\alpha(
Z)
=
\min_{\xi\in\R}
\E[\xi+\max(0,Z-\xi)/\alpha]$. 
Using the risk constraint \eqref{eq:avar:constraint} yields many advantages compared to other (e.g., chance-constrained) formulations, see \cite{LewBonalliPavoneRAL2024}.

The total cost $\ell_T(x_N)+\sum_{k=0}^{N-1}
\ell(x_k,u_k,u_{k+1})$ is also a random variable as it depends on the state trajectory, so we minimize its expected (average) value, with the objective of minimizing the average lap time. We formulate the risk-averse optimal control problem (\textbf{RA-OCP})
\begin{align*}
\min_{x,u}
\ 
&\E\left[\ell_T(x_N)+\textstyle\sum_{k=0}^{N-1}
\ell(x_k,u_k,u_{k+1})
\right]
&&
\hspace{0mm}\boxed{\scalebox{1.1}{\textbf{RA-OCP}}}
\\
\text{s.t.}
\ 
&f(x_{k+1},x_k,u_k,u_{k+1},\theta)=0, &&\hspace{-3mm}k\in[0,N-1],
\\
&(x_0,u_0)=(x_{\text{init}},u_{\text{init}}),
\\
&u_{\textrm{min}}\leq u_k\leq u_{\textrm{max}},
&&\hspace{-3mm}k\in[1,N],
\\
&\textrm{CVaR}_\alpha\left(
\max_{k=1,\dots,N}g_k(x_k)\right)\leq 0.
\end{align*}
\textbf{RA-OCP} encodes the problem of minimizing the average lap time (with additional regularization encoded by the cost term $\ell$) while bounding the tail probability of leaving the track, over the uncertain parameters $\theta\sim p_\theta$. 

\section{Sample-based reformulation and numerical resolution}
\textbf{RA-OCP} is a challenging problem to solve due to the uncertainty over the parameters $\theta$. Inspired by  \cite{LewBonalliPavoneRAL2024}, we compute approximate solutions to \textbf{RA-OCP} by solving a sample-based reformulation instead that depends on  $(1+M)$ samples $\theta^i$ of the parameters $\theta$. 

First, we select a nominal value $\bar\theta:=\theta^0$ for the parameters $\theta$ to parameterize a nominal state and control trajectory $(\bar{x},\bar{u}):=(x^0,u^0)$, which will be used to reduce uncertainty growth over the planning horizon via feedback. This trajectory satisfies the nominal dynamics
$$
f(\bar{x}_{k+1},\bar{x}_k,\bar{u}_k,\bar{u}_{k+1},\bar\theta)=0\ \forall k,
\
(\bar{x}_0,\bar{u}_0)=(x_{\text{init}},u_{\text{init}}).
$$

Second, we account for uncertainty over the parameters $\theta$ by parameterizing $M$ state and control trajectories  $(x^i,u^i)$ around the nominal trajectory $(\bar{x},\bar{u})$. We select %
$M$ %
samples $\theta^i$ of $\theta$ that define the state trajectories as %
\begin{align}\label{eq:states_particles}
&x_0^i=x_{\text{init}},
\ 
f(x_{k+1}^i,x_k^i,u_k^i,u_{k+1}^i,\theta^i)=0\,\   
\forall k,
\end{align}
and the closed-loop control trajectories as %
\begin{align}\label{eq:controls:closedloop}
&u_k^i=\bar{u}_k-K_k^i(x_k^i-\bar{x}_k)=u_k^0-K_k^i(x_k^i-x_k^0),
\end{align}
where 
the feedback gains $K_k^i$ are computed as the solution to a linear-quadratic regulator (LQR) problem~\protect{\cite[Chapter 12]{boyd1991linear}} around %
$(x_\textrm{ref},u_\textrm{ref})$ using the parameters $\theta^i$. %

Finally, we introduce the   optimization variables
\begin{align*}
X&=\left(\bar{x},x^1,\dots,x^M\right)
\in \R^{(1+M)Nn},
& \bar{u}&\in\R^{Nm},
\\
y&=\left(y^1,\dots,y^M\right)\in\R^M,
&
\xi&\in\R,
\end{align*}
where $y$ and $\xi$ are risk variables, 
the state and control dimensions are $(n,m)=(8,3)$, the sample size and horizon are $(M,N)=(10,32)$, 
and formulate the following sample-based approximation to \textbf{RA-OCP}   
\begin{align*}
\min_{X,\bar{u},y,\xi}
\ 
&\frac{1}{1+M}
\sum_{i=0}^M
\Big(
\ell_T(x_N^i)+\textstyle\sum_{k=0}^{N-1}
\ell(x_k^i,\bar{u}_k,\bar{u}_{k+1})
\Big)
\\
\text{s.t.}
\ \ 
&f(x_{k+1}^i,x_k^i,u_k^i,u_{k+1}^i,\theta^i)=0, 
&&\hspace{-28mm}
\begin{cases}
    k\in[0,N-1],
    \\
    i\in[0,M],
\end{cases}
\\
&u_k^i=\bar{u}_k-K_k^i(x_k^i-\bar{x}_k)
&&\hspace{-28mm}
\begin{cases}
    k\in[0,N],
    \\
    i\in[0,M],
\end{cases}
\\
&(x_0^i,\bar{u}_0)=(x_{\text{init}},u_{\text{init}}),
&&\hspace{-25mm}i\in[0,M],
\\
&u_{\textrm{min}}\leq \bar{u}_k\leq u_{\textrm{max}},
&&\hspace{-25mm}k\in[1,N],
\\
&e_{\textrm{min}}\leq \bar{e}_k\leq e_{\textrm{max}},
&&\hspace{-25mm}k\in[1,N],
\\
&\begin{cases}
(M\alpha)\xi+\sum_{i=1}^My^i\leq 0 
\\
0\leq y^i
&
\hspace{-4mm}
i\in[1,M],
\\
g_k(x^i_k)-\xi\leq y^i,
\ 
&%
\hspace{-4mm}
i\in[1,M],\,
k\in[1,N].
\end{cases}
\end{align*}

\begin{remark}[On closed-loop uncertainty propagation]
Using a feedback law in the optimization problem \textbf{RA-OCP}  to define closed-loop trajectories  as in \eqref{eq:states_particles}-\eqref{eq:controls:closedloop} is a standard approach  \cite{Liniger2017,Quirynen2021,Lew2020} to prevent uncertainty to grow unbounded over time and make the optimization infeasible. It also allows approximately accounting for closed-loop feedback of the receding horizon MPC scheme. 
We do not enforce input constraints for the controls $u^i$ because solutions are already constrained by the tire friction limits, 
although one could %
include closed-loop input constraints in the risk constraint \eqref{eq:avar:constraint} or saturate the controls $u^i$ within the dynamics \cite{Leparoux2024}.
\end{remark}

\subsection*{Efficient numerical resolution of the sample-based problem: leveraging GPU parallelization and sparsity}
Efficiently solving the sample-based reformulation of \textbf{RA-OCP} requires special care due to the large size of the problem. To achieve fast replanning, we leverage two observations. First, the problem is sparse, e.g., the particle $x^i$ does not explicitly depend on $x^{i+1}$.   Second, since we parameterize the state trajectory $x^i$ for each particle, the cost terms and constraints can be evaluated in parallel over both particles and time on a GPU.

Thus, we decide to solve the sample-based reformulation of \textbf{RA-OCP} via a sequential quadratic programming (SQP) method with a linesearch \cite[Chapter 18]{Nocedal2006}. We evaluate the cost, constraints, and their gradients in Python using \texttt{Jax} \cite{jax2018github}, so we can easily evaluate the quadratic programs (QPs) at each SQP step on a GPU and take advantage of the parallel structure of the sample-based reformulation. %
The problem data is then moved to the CPU and the QPs are solved using \texttt{OSQP} \cite{Stellato2020}. Even though the QPs involve many variables, they are quickly solved thanks to the sparsity of the QPs that \texttt{OSQP} leverages internally. For receding horizon MPC, we  only perform one SQP  iteration at each timestep \cite{Diehl2005}.

\begin{remark}[Alternatives and lessons learned]\label{remark:alternatives_lessons_learned}
\hspace{1mm}The \\nominal trajectory $\bar{x}$ could be removed from the optimization variables by defining $\bar{x} =x^0=\frac{1}{M}\sum_{i=1}^Mx^i$. 
However, the resulting formulation densely couples the particles $x^i$ via the feedback controls in \eqref{eq:controls:closedloop} and is thus slow to solve numerically. Parameterizing the nominal trajectory greatly increases the sparsity of the problem. 

We tried only optimizing over the control inputs $u$ and computing states $x^i(u)$ for each particle via an explicit integration scheme %
as in \cite{LewBonalliPavoneRAL2024}, which we found to be %
numerically sensitive to the choice of initial guess. Parameterizing the particles $x^i$ and using an implicit integrator instead improves robustness and enables parallelizing computations over times $k$ and samples $i$.

\end{remark}

\section{Problem Setup and Trajectory Analysis}\label{sec:results:simulation}
We consider an oval racing track shown in Figure \ref{fig:hardware:racing_wet:aerial}. 
The reference trajectory $(x_{\textrm{ref}},u_{\textrm{ref}})$ is the optimal racing line computed offline for nominal parameters $\bar\theta$ using the method in \cite{Dallas2023}.  In this section, we study trajectories solving \textbf{OCP} and \textbf{RA-OCP} for different values of the parameters $\theta$. For visualization purposes, we use $M=5$ samples of the parameters, shown in Table \ref{table:parameters}.

\begin{figure}[!t]
    \centering
        \includegraphics[width=1.0\linewidth]{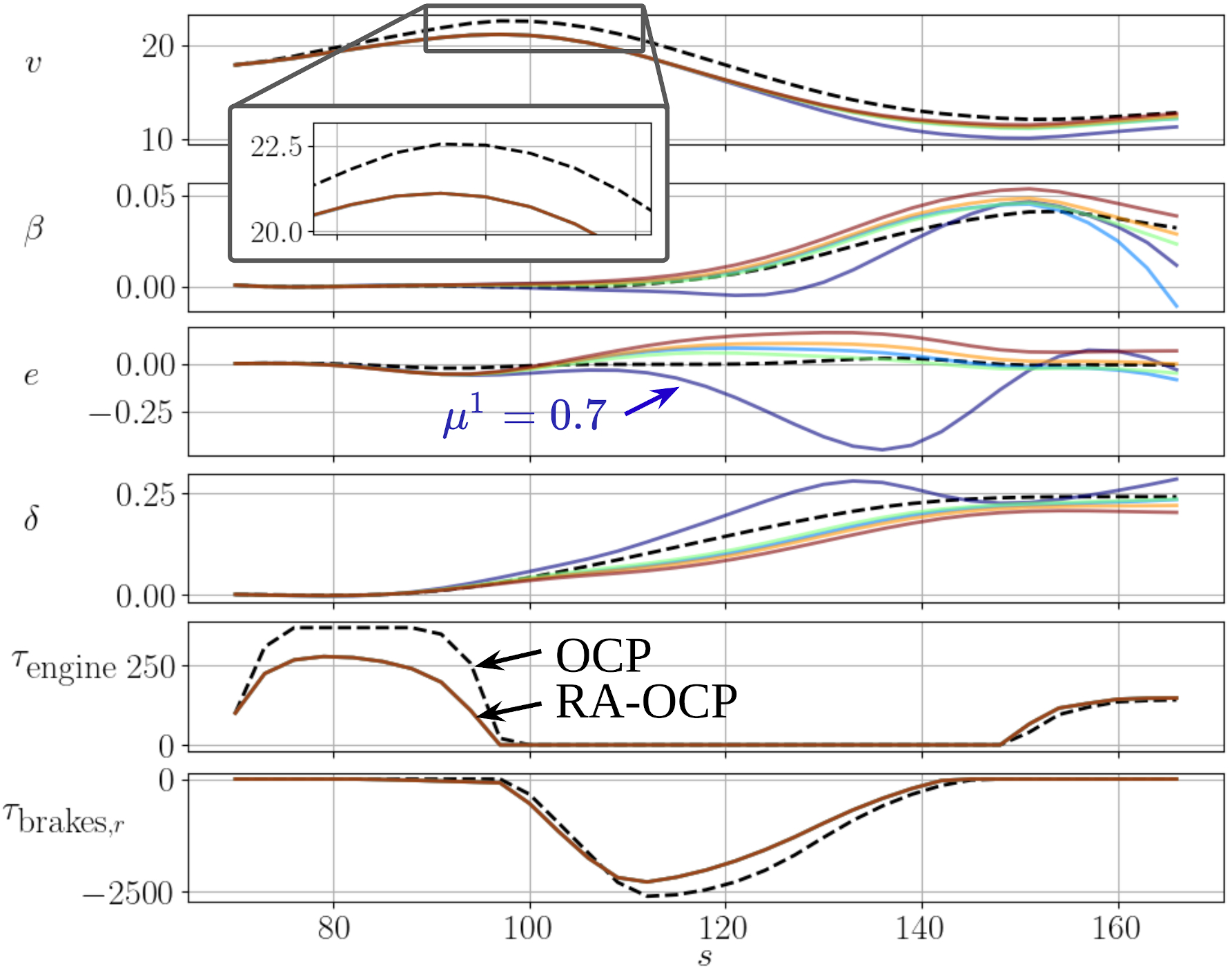}
        \\[-2mm]%
        \vspace{-5mm}
\includegraphics[width=0.5\linewidth]{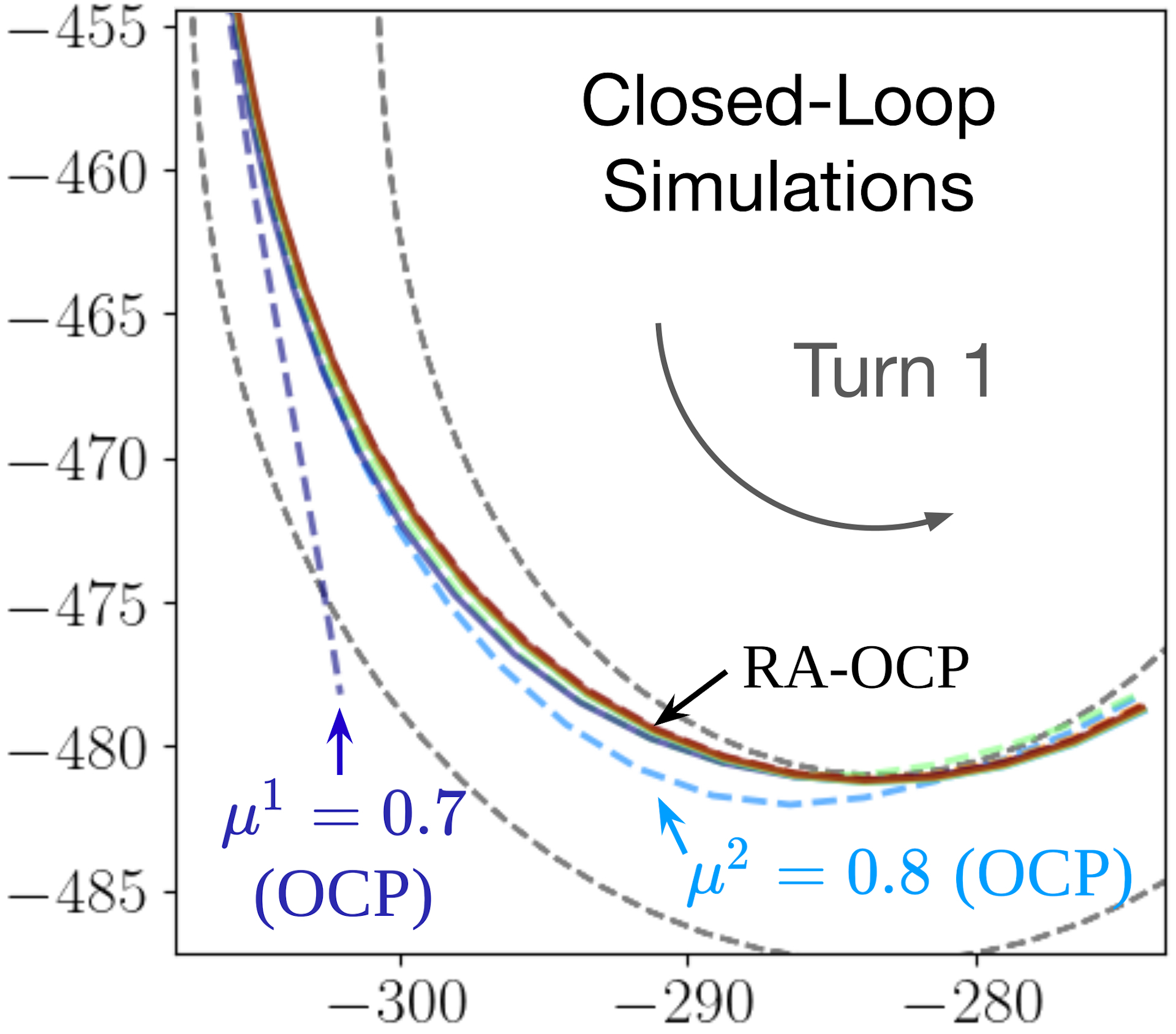}
    \hspace{0.065\linewidth}
        \begin{minipage}[c][1\linewidth]{0.4\linewidth}%
        \vspace{-34mm}
        \begin{center} 
        \includegraphics[width=1\linewidth]{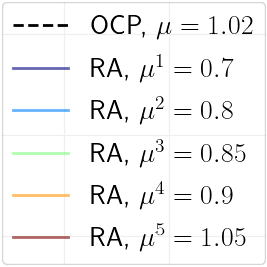}
        \end{center}
        \end{minipage}
        \\[-40mm]
    \caption{
    Top: Nominal vs risk-averse  solutions. 
    Bottom: 
    Closed-loop trajectories for different parameters $\theta$ corresponding to solutions to \textbf{OCP} (dashed lines) and \textbf{RA-OCP} (solid lines).
    \vspace{-4mm}
    }%
    \label{fig:simulation:trajopt}
\end{figure}

\textit{a) How different are risk-averse solutions compared to solutions to \textbf{OCP} with nominal tire parameters $\bar\theta$?} 
We present trajectories solving \textbf{OCP} and \textbf{RA-OCP} in Figure \ref{fig:simulation:trajopt}. The risk-averse solution applies less engine torque and braking forces to account for potentially reduced tire forces (see Figure \ref{fig:tire_forces}) and brakes earlier than the solution to \textbf{OCP}. This results in a lower velocity in the turn. We also plot closed-loop trajectories $x^i$ satisfying \eqref{eq:states_particles} for \textbf{OCP} for the different parameters $\theta^i$. The solution to \textbf{OCP} may drive too fast to complete the turn: the vehicle may slide out of the track if the tire friction parameters $\mu_f$ and $\mu_r$ take lower values. In contrast, solutions to \textbf{RA-OCP} account for different tire parameter values to ensure robust handling of the vehicle in the turn. 

\begingroup
\def\arraystretch{1.1}
\setlength\tabcolsep{0.68mm}
\begin{table}[!t]
\caption{Tire parameters $\theta^i$ defining \textbf{OCP} and \textbf{RA-OCP} for trajectory analysis (Sec.\ref{sec:results:simulation}) (top) and MPC (Sec.\ref{sec:results:hardware}) (bottom).
\vspace*{-5mm}
}
\label{table:parameters}
\centering\begin{tabular}{c|c|c|c|c|c|c}
\toprule
\hspace{1mm}
\hspace{5mm}Trajectory Analysis\hspace{5mm}
& $\bar\theta{=}\theta^0$ 
& $\theta^1$ & $\theta^2$ & $\theta^3$ & $\theta^4$
& $\theta^5$
\\ 
\midrule
$\mu_f$ & $1.02$ & $0.70$ & $0.80$ & $0.85$ & $0.90$ & $1.05$
\\ 
$\mu_r$ & $1.08$ &  $0.70$ & $0.80$ & $0.85$ & $0.90$ & $1.05$
\\ 
$10^{-3}C_f$ & $115$ & $80$ & $140$ & $120$ & $140$ & $160$
\\ 
$10^{-3}C_r$ & $280$ & $240$ & $320$ & $280$ & $300$ & $320$
\\\bottomrule
\end{tabular}
\\[1mm]

\setlength\tabcolsep{0.3mm}
\centering\begin{tabular}{c|c|c|c|c|c|c|c|c|c|c|c}
\toprule
MPC
& $\bar\theta{=}\theta^0$ 
& $\theta^1$ & $\theta^2$ & $\theta^3$ & $\theta^4$
& $\theta^5$ & $\theta^6$ & $\theta^7$ & $\theta^8$
& $\theta^9$ & $\theta^{10}$
\\ \midrule
$\mu_f$ & $1.02$ & $0.75$ & $0.74$ & $1.05$ & $0.92$ & $0.93$ & $0.94$ & $0.95$ & $0.96$ & $0.97$ & $0.98$
\\ 
$\mu_r$ & $1.08$ & $0.75$ & $0.74$ & $1.05$ & $0.92$ & $0.93$ & $0.94$ & $0.95$ & $0.96$ & $0.97$ & $0.98$
\\ 
$10^{-3}C_f$ & $115$ & $90$ & $130$ & $92$ & $125$ & $95$ & $121$ & $95$ & $110$ & $97.5$ & $115$
\\ 
$10^{-3}C_r$ & $280$ & $240$ & $320$ & $250$ & $300$ & $280$ & $290$ & $290$ & $310$ & $280$ & $290$
\\
\bottomrule
\end{tabular}
\vspace{-6mm}
\end{table}
\endgroup

\textit{b) Does solving \textbf{OCP} with   low-friction tire parameters give robust performance?} 
\begin{figure}[!t]
    \centering
        \includegraphics[width=1.0\linewidth]{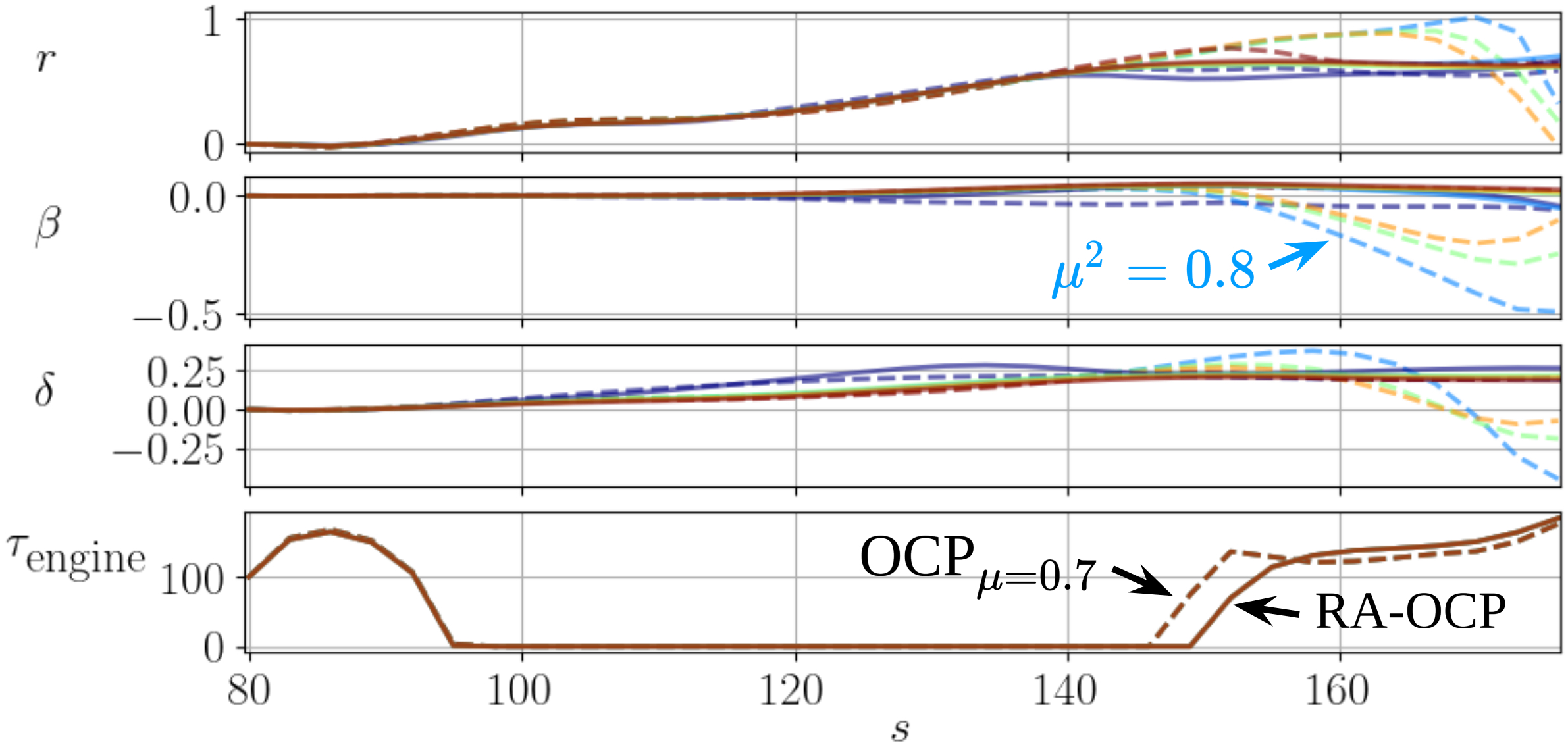}
        \\[-1.1mm]
        \includegraphics[width=1\linewidth]{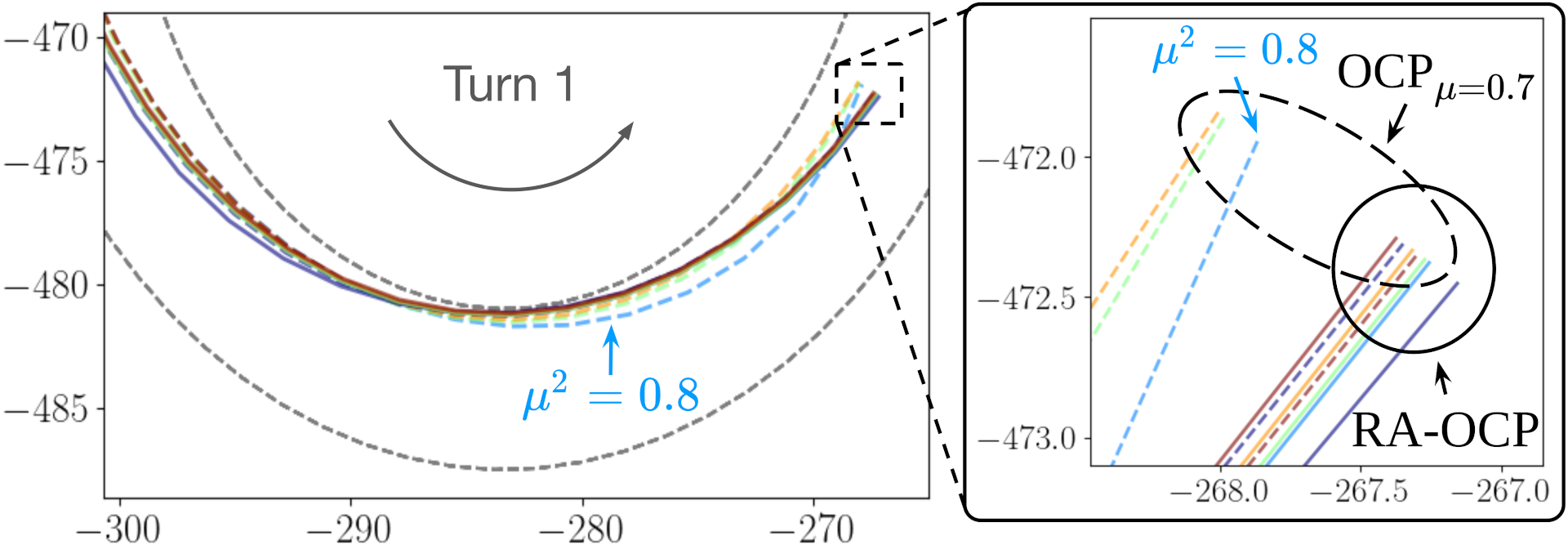}
    \caption{Closed-loop trajectories for different parameters $\theta$ corresponding to solutions to \textbf{OCP} with low friction values $\mu_f=\mu_r=0.7$ (dashed lines), and \textbf{RA-OCP} (solid lines). 
    \vspace{-7mm}
    }
    \label{fig:simulation:trajopt:riskaverse_vs_ocp_low_mu}
\end{figure} 
We solve \textbf{OCP} assuming that $\bar\theta=\theta^1$ (corresponding to low-friction tire parameters $\mu_f=\mu_r=0.7$) and present results in Figure \ref{fig:simulation:trajopt:riskaverse_vs_ocp_low_mu}. The solution to \textbf{OCP} is now closer to the solution to \textbf{RA-OCP} and yields better performance over different tire parameter values $\theta^i$ than if assuming larger-friction tire parameters. %
However, %
the performance remains poor for different tire parameter values, e.g., exhibiting large sideslip ($\beta$) values for $\mu_f=0.8$ that may lead to suboptimal closed-loop performance. Overall, these results show that accounting for different tire parameters by solving \textbf{RA-OCP} gives additional robustness compared to only planning with low-friction tire parameters. 

\section{Hardware Results}\label{sec:results:hardware}
We validate the risk-averse controller on a 2019 Lexus LC 500. State estimates come from an OxTS Inertial Navigation System~\cite{oxts2022}. The optimization problems are solved %
on an on-board computer with an  Intel Xeon E-2278GE CPU @3.30GHz and an NVIDIA GeForce RTX 3070 GPU. Further details about the vehicle are in \cite{Dallas2023}. 

We compare %
the proposed risk-averse MPC approach solving \textbf{RA-OCP} with a nominal MPC solving \textbf{OCP} with the parameters $\theta=\bar\theta$. For all experiments, we fix the parameter samples $\theta^i$ in \textbf{OCP} and \textbf{RA-OCP} to the values in Table \ref{table:parameters} to cover the parameter space. Keeping these parameter values constant in time 
also removes randomness in the algorithm. The nominal MPC runs at about $100$Hz. The risk-averse MPC runs 
at $20$Hz 
(with mean $22$Hz and standard deviation $1.5$Hz). %
These update rates are sufficient to reliably control the vehicle.

\begin{figure}[!t]
    \centering
    \includegraphics[width=1.0\linewidth]{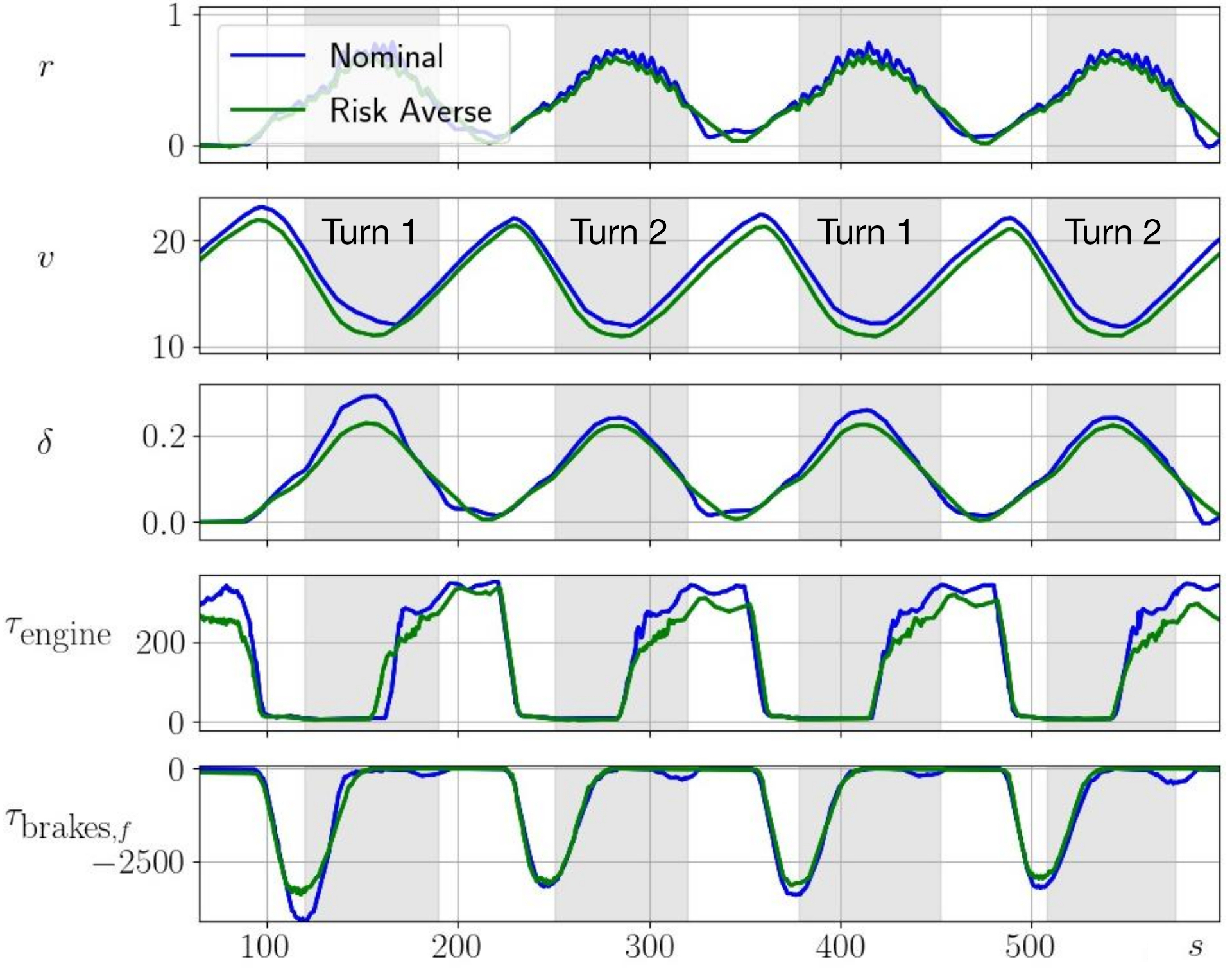}
    \vspace{-17pt} 
    
    \caption{Racing results in nominal dry conditions. A single run with the nominal (blue) and risk-averse MPC (green).
\vspace{-3mm}
}
    \label{fig:hardware:racing_dry}
\end{figure}

\textit{a) Racing in dry conditions}: 
We first validate the controllers in dry conditions and report results in Figure \ref{fig:hardware:racing_dry}. 
Nominal MPC (solving \textbf{OCP} with the parameters $\bar\theta$) applies larger engine and braking torques and drives slightly faster than risk-averse MPC (solving \textbf{RA-OCP}), with a top speed difference of $1\textrm{m/s}$. Nominal MPC causes slight yaw rate oscillations while turning due to tire saturation. %
Overall, nominal MPC slightly outperforms risk-averse MPC in such dry conditions, which is expected behavior given that risk-averse MPC optimizes the expected final time over a wider range of parameters including lower friction values $(\mu_f,\mu_r)$.
We also tested nominal MPC planning with low-friction parameters ($\mu_f=\mu_r=0.7$). This approach consistently cuts corners in the turns due to model mismatch and is thus unable to race in dry conditions.

\textit{b) Racing through a patch of water}: 
We test again the %
controllers on a dry track with a water patch in turn $2$ (see Figure \ref{fig:hardware:racing_wet:aerial}). Such conditions may represent a track that is slowly drying up after rain, resulting in a leftover patch of water. We run the controllers three times each. For repeatability, we alternate between nominal and risk-averse MPC and add water after each individual run.

\begin{figure}[!t]
    \centering
    \includegraphics[width=1.0\linewidth]{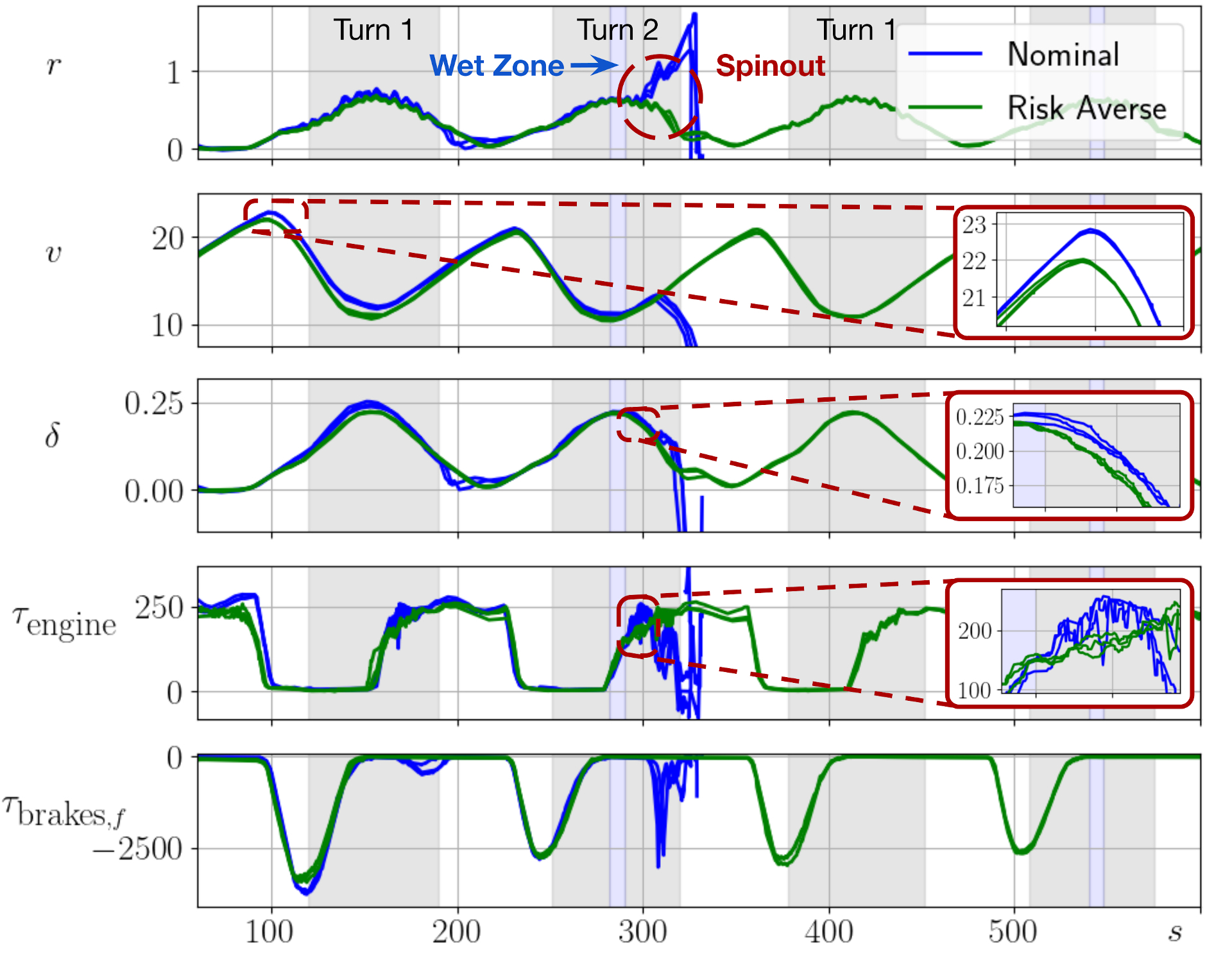}
\vspace{-6mm}
\caption{Racing results with a wet area (see also Figure \ref{fig:hardware:racing_wet:aerial}). Three runs of the nominal (blue) and risk-averse MPC (green), with close ups at peak velocity and after the wet zone in turn $2$.
\vspace{-5mm}
}
    \label{fig:hardware:racing_wet}
\end{figure}

Results are reported in Figure \ref{fig:hardware:racing_wet}. %
After driving through the wet area when exiting turn $2$, nominal MPC consistently causes a spin out, due to applying engine torques  that are too large while oversteering. Indeed, tires are wet and thus cannot apply the planned tire forces, which results in a spin out. In contrast, risk-averse MPC applies slightly smaller engine torques when exiting each turn and is thus consistently able to drive through the wet area and complete the lap. 
While more robust behavior may be obtained with nominal MPC by reducing the minimum-time cost $\ell_T$ compared to the regularizer $\ell$, such a strategy may require additional hyperparameter tuning and lead to suboptimality and slower lap times.


\section{Conclusion}
We presented a risk-averse MPC framework for racing and showed that accounting for different tire parameters provides a natural avenue for infusing robustness into MPC. By leveraging a particular sample-based risk-averse formulation, our method accurately accounts for uncertain nonlinear tire dynamics and is amenable to online replanning in MPC. 
In future work, we plan on investigating improvements to the SQP solver (e.g., by developing specialized methods inspired from \cite{Plancher2020,Schubiger2020,bishop_relu-qp_2023,adabag2024mpcgpu} to solve \textbf{RA-OCP} entirely on the GPU) to unlock faster replanning and using larger sample sizes,  
and interfacing with perception to account for state uncertainty and parameters that vary over time and space.

\section*{Acknowledgments}
\noindent 
We thank Phung Nguyen, Steven Goldine, and William Kettle for their support with the test platform and experiments, and Jenna Lee for helping with data visualization.

\bibliographystyle{IEEEtran}
\bibliography{main}

\end{document}